\documentclass[11pt]{article}

\date{}

\usepackage[utf8]{inputenc} 
\usepackage[T1]{fontenc}    
\usepackage{hyperref}       
\usepackage{url}            
\usepackage{booktabs}       
\usepackage{amsfonts}       
\usepackage{nicefrac}       
\usepackage{microtype}      
\usepackage{xspace}
\usepackage{natbib}

\usepackage{comment}
\usepackage{amsmath}
\usepackage{amsthm}
\usepackage{graphicx}
\usepackage{rotating}

\usepackage{amssymb}
\usepackage{mathtools}
\usepackage{xcolor}
\usepackage{bbm}
\usepackage{algorithm}

\title{Stability and Identification of Random Asynchronous Linear Time-Invariant Systems}
\makeatletter

\usepackage{natbib}

\usepackage{wrapfig}
\usepackage{amsthm}
\usepackage{enumitem}
\usepackage{subfig}
\usepackage[export]{adjustbox}
\usepackage{algorithmicx}
\usepackage[noend]{algpseudocode}

\newcommand{\ee}{\mbox{\boldmath$e$}\xspace}

\newcommand{\uu}{\mbox{$\bf u$}}

\newcommand{\xx}{\mbox{$\bf x$}}

\newcommand{\ww}{\mbox{$\bf w$}}

\newcommand{\AABOLD}{\mbox{$\bf A$}}
\newcommand{\BB}{\mbox{$\bf B$}}
\newcommand{\CC}{\mbox{$\bf C$}}

\newcommand{\HH}{\mbox{$\bf H$}}
\newcommand{\II}{\mbox{$\bf I$}}
\newcommand{\JJ}{\mbox{$\bf J$}}

\newcommand{\MM}{\mbox{\boldmath$M$}\xspace}

\newcommand{\TT}{\mbox{$\bf T$}}

\newcommand{\XX}{\mbox{$\bf X$}}

\newcommand{\bzero}{\mbox{\boldmath$0$}\xspace}

\usepackage{ amsmath, psfrag, xspace, mathabx, dsfont, enumitem}

\newcommand{\fro}{\textnormal{F}}

\newcommand{\inCorr}{\mbox{$\bf U$}}
\newcommand{\pr}{p}

\newcommand{\randA}{\widebar{\opM}}
\newcommand{\randB}{\widebar{\BB}}
\newcommand{\randH}{\widebar{\HH}}
\newcommand{\edlyap}{\texttt{ly}\xspace}
\newcommand{\vecMap}{\mbox{$\bf S$}}
\newcommand{\coefM}{\mbox{$\bf T$}}

\newcommand{\markovMat}{\mbox{$\bf G$}}
\newcommand{\randmarkovMat}{\widebar{\markovMat}}

\newcommand{\xCor}{\mbox{\boldmath$\Gamma$}}

\newcommand{\dimIn}{d_u}
\newcommand{\dimSta}{d_x}

\newtheorem{myLemma}{Lemma}

\DeclareMathOperator{\tr}{tr}

\DeclareMathOperator{\vect}{vec}

\newcommand{\opM}{\AABOLD}
\newcommand{\selMat}{\mbox{$\bf E$}}
\newcommand{\expc}{\mathbb{E}}
\newcommand{\prob}{\mathbb{P}}

\author{%
  Sahin Lale\thanks{equal contribution} , Oguzhan Teke\footnotemark[1] , Babak Hassibi, Anima Anandkumar\\
  California Institute of Technology\\
  \texttt{\{alale,oteke,hassibi,anima\}@caltech.edu} \\
}

\begin{document}

\maketitle

\begin{abstract}%
In many computational tasks and dynamical systems, asynchrony and randomization are naturally present and have been considered as ways to increase the speed and reduce the cost of computation while compromising the accuracy and convergence rate. In this work, we show the additional benefits of randomization and asynchrony on the stability of linear dynamical systems. We introduce a natural model for random asynchronous linear time-invariant (LTI) systems which generalizes the standard (synchronous) LTI systems. In this model, each state variable is updated randomly and asynchronously with some probability according to the underlying system dynamics. We examine how the mean-square stability of random asynchronous LTI systems vary with respect to randomization and asynchrony. Surprisingly, we show that the stability of random asynchronous LTI systems does not imply or is not implied by the stability of the synchronous variant of the system and an unstable synchronous system can be stabilized via randomization and/or asynchrony. We further study a special case of the introduced model, namely randomized LTI systems, where each state element is updated randomly with some fixed but unknown probability. 
We consider the problem of system identification of unknown randomized LTI systems using the precise characterization of mean-square stability via extended Lyapunov equation. For unknown randomized LTI systems, we propose a systematic identification method to recover the underlying dynamics. Given a single input/output trajectory, our method estimates the model parameters that govern the system dynamics, the update probability of state variables, and the noise covariance using the correlation matrices of collected data and the extended Lyapunov equation. Finally, we empirically demonstrate that the proposed method consistently recovers the underlying system dynamics with optimal rate.
\end{abstract}

\section{Introduction}

\noindent \textbf{Randomization, asynchrony and stability:} Randomization and asynchrony are crucial to many computational tasks that involve a large number of agents working cooperatively with each other \citep{fagnani2008randomized,async_mapReduce}. They allow speed ups and cost reduction in many artificial and biological processes by removing the synchronization time, relaxing the communication bottlenecks, minimizing the cost of cooperation, and increasing efficiency. For example, large scale control systems with multiple sensors adopt random asynchronous updates from its sensors due to power saving and difficulty of synchronization \citep{hespanha2007survey}. Similarly, asynchrony and randomization are central elements in the dynamical systems of biological neural networks \citep{singer1999neuronal,uhlhaas2009neural}. In various studies, e.g. \citep{visual,auditory}, researchers have found out that the synchrony/asynchrony balance phenomenon is ubiquitous in cortical networks. They show the existence of a delicate equilibrium between synchrony and asynchrony of neural firings in many cognitive tasks and regions of the brain such as visual, auditory, and memory maintenance to obtain stable dynamics during computations. Any disturbance to this natural equilibrium may result in neurological disorders \citep{brain_disOrd}.

\noindent \paragraph{Random asynchronous LTI systems:} In modeling dynamical systems, linear dynamical systems (LDS) are the main focus due to their simplicity and ability to capture the crux of the problem and give insights on more challenging settings. Consider the following state-space model of LTI systems:
\begin{equation} \label{sync_system}
    \xx_{t+1} = \AABOLD \, \xx_{t} + \BB \, \uu_t+ \ww_t,
\end{equation}
where $\xx_t, \uu_t$ and $\ww_t$ are the state, input, and noise vectors respectively. This model can be referred to as ``synchronous LTI system" since all state elements are updated in every time-step using the most recent information on all state elements. In this work, we study a random asynchronous variant of this LTI system where the states evolve randomly and asynchronously. The model studied here generalizes \eqref{sync_system} in two different directions:

1) The state elements get updated randomly in every iteration. If a state element gets updated, it follows the state dynamics. Otherwise, its value remains the same.

2) When a state element is being updated, it may use out-dated information regarding the other state elements. We assume that the delay in information flow is also probabilistic.


More precisely, given a node update probability ${0 < \pr \leq 1}$, we consider the following \emph{random and asynchronous} state-space model for all state variables $i$ at each time step $t$:
\begin{align} \label{eq:asyncStateRec}
(\xx_{t\textrm{+}1})_i &= \begin{cases}
\displaystyle \sum\nolimits_{j=1}^n A_{i,j} \, (\xx_{t-k_{i,j}})_j + (\BB \, \uu_t + \ww_{t})_i, &\quad \textnormal{w.p.} \qquad \pr, \\
(\xx_{t} + \ww_t)_i, &\quad \textnormal{w.p.} \qquad 1 - \pr,
\end{cases} 
\end{align}
where ${\ww_{t}}$ denotes the noise component for the state elements. More importantly, ${k_{i,j} \geq 0}$ denotes the delay in information observed by the ${i^{th}}$ element regarding the ${j^{th}}$ element. So, state variables are allowed to observe different amount of delay regarding other state variables. In our model, we will consider \emph{random and independent delays} with the following distribution:
\begin{equation}
\prob[ k_{i,j} = \tau] = \begin{cases} q \, (1-q)^{\tau}, & \tau = \{0, \ldots, h-1 \}, \\
(1-q)^{h}, & \tau = h,
\end{cases}
\end{equation}
for some fixed delay probability $0 < q \leq 1$ and finite $h$. So, higher values of ${q}$ implies lower amount of delay in information flow. In summary, in every time-step, each state element is updated with probability ${\pr}$ using linear dynamics based on the most recent data available from other state variables, or its value remains the same (up to an input noise) with probability ${1-\pr}$.

The model \eqref{eq:asyncStateRec} captures the random asynchrony of large-scale LDS (e.g. networked control systems, social networks, and biological networks), where each state variable could be considered as a sensor/node. At any time step the update on the node happens randomly. The node may not have the most recent data from the other nodes, and it updates its state based on the available (possibly outdated) information. It is possible to extend this model in such a way that each state variable has a different update probability, a different delay probability or updated and non-updated state elements have different noise components. However, for the sake of simplicity, we assume that all the state variables are updated with the same probability, the same delay scheme and the same noise characteristics. Notice that the model \eqref{eq:asyncStateRec} reduces to the standard \textit{synchronous} state-space model (\ref{sync_system}) when the probabilities are selected as $\pr=1$ and $q =1$. As a result, one can consider \eqref{eq:asyncStateRec} as a random asynchronous extension of synchronous LTI systems that are studied in the last century.

\noindent \paragraph{Mean-square stability:} In stochastic systems, mean-square stability is one of the most important notions of stability. For a system with a fixed-point, mean-square stability means that the system converges to its' fixed point asymptotically in mean-square sense. For a noisy system as in (\ref{sync_system}) or (\ref{eq:asyncStateRec}), it implies that the covariance matrix of state vector stays finite and converges to the solution of Lyapunov equation of the system, \textit{i.e.}, steady-state covariance matrix. However, the Lyapunov equation and mean-square stability characterization of (\ref{sync_system}) do not hold for (\ref{eq:asyncStateRec}).

\noindent \paragraph{System identification:} In modeling the dynamics of a system, the underlying system is usually unknown and only a sequence of inputs and outputs is available. This raises the system identification problem which aims to recover the parameters that govern the dynamics from the data collected. The classical and recent system identification methods mainly focus on LDS and consider stable synchronous LTI systems or switching linear systems \citep{ljung1996subspace,lale2020logarithmic,ozay2015set,sarkar2019nonparametric}. 
For switching linear systems, the system identification methods require the knowledge of the order of switched systems, otherwise they become computationally intractable and sample inefficient due to exponential dimension dependency \citep{lauer2018hybrid}. Thus, they have limited applicability to the large scale practical random asynchronous LTI systems. This highlights the necessity of a careful and systematic approach in deriving stability conditions and system identification framework of random asynchronous LTI systems.
  
\noindent \paragraph{Our Contributions:} In this work, we study the stability characterization and the system identification problem for random asynchronous LTI systems \eqref{eq:asyncStateRec}. We have the following contributions:

\begin{enumerate}[wide, labelindent=0pt,font=\bfseries]
    \item \textbf{Empirical Study of Stability of Random Asynchronous LTI Systems:} We empirically show that the model \eqref{eq:asyncStateRec}, which governs the dynamics of random asynchronous updates in large scale LTI systems, brings novel challenges and opportunities in terms of stabilization of LDS. We empirically demonstrate that the mean-square stability of random asynchronous LTI systems has a delicate dependency on the probabilities $p$ and $q$. Furthermore, the stability of synchronous LDS does not imply stability of asynchronous LDS, and vice versa. For a randomly generated system, by changing the asynchrony or the update probability in the system, we show that unstable synchronous LTI systems may be stabilized or stable synchronous LTI systems may have unstable dynamics.
    \item \textbf{Theoretical Study of Stability of Randomized LTI Systems:} We consider the mean-square stability characterization of randomized LTI systems, which are special cases of random asynchronous LTI systems with $q = 1$. We show that this setting corresponds to the model introduced in \citet{teke_rand_async}, and the stability is governed by an extended Lyapunov equation. Relating the extended Lyapunov equation to standard Lyapunov equation of synchronous LTI system, we discuss the precise characterization of the mean-square stability of randomized LTI systems.
    
    \item \textbf{Novel System Identification for Unknown Stable Randomized LTI Systems:} We propose a novel method to recover the impulse response of the ``average system", as well as the true underlying system parameters, the update probability $p$, noise, and input covariances for unknown stable randomized LTI systems. In order to achieve this, we first visit the well-known central limit theorem for Markov chains and solve a least-squares problem to obtain an estimate of the average system parameters of the underlying system. Then, we propose an optimization problem to estimate the update probability and input/noise covariances that optimally satisfy the extended Lyapunov equation for the estimated average system parameters. By solving the optimization problem analytically, we present a closed-form expression for an estimate of the update probability and input/noise covariances for the system. The underlying true system parameters that govern the dynamics are ultimately recovered via the estimates of the update probability and the average system parameters.

    \item \textbf{Empirical Study of Novel System Identification Framework:} 
    We empirically demonstrate the performance of the novel method on a simulated randomized LTI system and show that our proposed method reliably and efficiently recovers the underlying dynamics with the optimal rate. This shows that the presented model and the system identification framework provide a more realistic and computationally efficient alternative to the general switching linear systems in analyzing input/output data collected from an LDS that has a fixed network structure with random asynchronous updates.
\end{enumerate}





\section{Related Work}

In modeling stochastic or varying dynamics like random asynchronous LTI systems, there has been a strong interest in switching linear systems/Markov jump systems \citep{tanner2007flocking,sun2006switched,markovJumpSystems,zhang2009stability,zhang2005new,olfati2004consensus}, in which state variables evolve according to a randomly selected model among all possible models. Although randomized linear models can be studied under this framework, the number of possible models becomes exponential in the number of state variables, making this approach prohibitive for large-scale systems. Moreover, the connections between the nodes in randomized systems are mostly fixed without any switching between different systems. Having these connections on or off is the main cause of randomization within the system. Thus, in these dynamical systems, the underlying system is time-invariant while the active interaction within the system is time-varying. Prior works that adopt switching linear systems fail to capture this nature of random asynchronous LTI systems. 

In addition to the switching systems viewpoint, statistical behavior of \eqref{eq:asyncStateRec} can be also studied from the product of random matrices perspective~\citep{infProdNonnegMat, joint_spec_rad_book, convInfiniteProd, prodIntervalMatrices, revist_random, arock}. However, these frameworks usually come with additional constraints on the system matrix ${\opM}$, e.g., \citet{infProdNonnegMat} requires ${\opM}$ to be element-wise nonnegative and \citet{revist_random} requires ${\opM}$ to be positive definite. Similarly, approaches based on \emph{joint spectral radius} are too restrictive to reveal the effect of randomization \citep{joint_spec_rad_book}.

When the input is constant, i.e., $\uu_{t} = \uu \;\; \forall t$, \eqref{eq:asyncStateRec} reduces to asynchronous \emph{fixed-point} iterations, whose non-random variants are well-studied in the literature \citep{chazan_relax, baudet_async}. These studies show that in the case of linear updates, convergence of asynchronous fixed-point iterations is more restrictive than stability of $\opM$. However, it should be noted that these classical studies consider the convergence of the worst case asynchronous behavior, whereas this study focuses on the statistical behavior and shows that stability of $\opM$ is not required for mean-square convergence.

\section{Preliminaries} \label{sec:randModel}

\paragraph{Notation:} For matrix $\MM$, $\| \MM \|_2$ is its spectral norm, $\| \mathbf{M}\|_F$ is its Frobenius norm, $\mathbf{M}^\top$ is its transpose. For square matrices, $\tr(\cdot)$ is the trace and $\rho(\cdot)$ is the spectral radius. $\delta(t)$ denotes the unit impulse function. The Kronecker product is denoted as $\otimes $ and $\odot$ denotes the Hadamard product.

 \paragraph{Setting:} For the given system in \eqref{eq:asyncStateRec}, $\xx_t, \ww_t \in \mathbb{R}^{\dimSta}$, $u_t \in \mathbb{R}^{\dimIn}$, and the system matrices $\opM$ and $\BB$ have appropriate dimensions accordingly. Note that this work considers the problem in real domain for a simpler presentation. Nevertheless, the results can be easily extended to complex domain with proper conjugation operations. 
Without loss of generality, we assume that $\xx_0 = \bzero$. Furthermore, we assume that the input and noise vectors are zero mean and have unknown variance, \textit{i.e.}, 
\begin{align}
\expc[\uu_t] = \expc[\ww_t] = \bzero, \enskip \expc[\uu_t^{} \, \uu_k^{\top}] =  \inCorr , \enskip \expc[\ww_t^{} \, \ww_k^{\top}] = \delta(t-k) \, \, \sigma_w^2 \, \II_{\dimSta} \label{noise_characteristic}
\end{align}
for some \textit{unknown} $\inCorr \succeq 0 $ and $\sigma_w^2>0$. Note that we do not have any further assumption on the input and noise characteristics.

\section{Stability of Random Asynchronous LTI Systems} \label{sec:RA-LTI}

Since the random asynchronous LTI systems have stochastic behavior, the stability of the state vector ${\xx_k}$ should be considered statistically as well. Therefore, we first numerically study the mean-square stability of (\ref{eq:asyncStateRec}). Note that for this setting, the mean-square stability also implies almost-sure stability.

\begin{figure}[t]
  \centering
  \begin{minipage}[b]{0.48\textwidth}
  \centering
    \includegraphics[width=\textwidth]{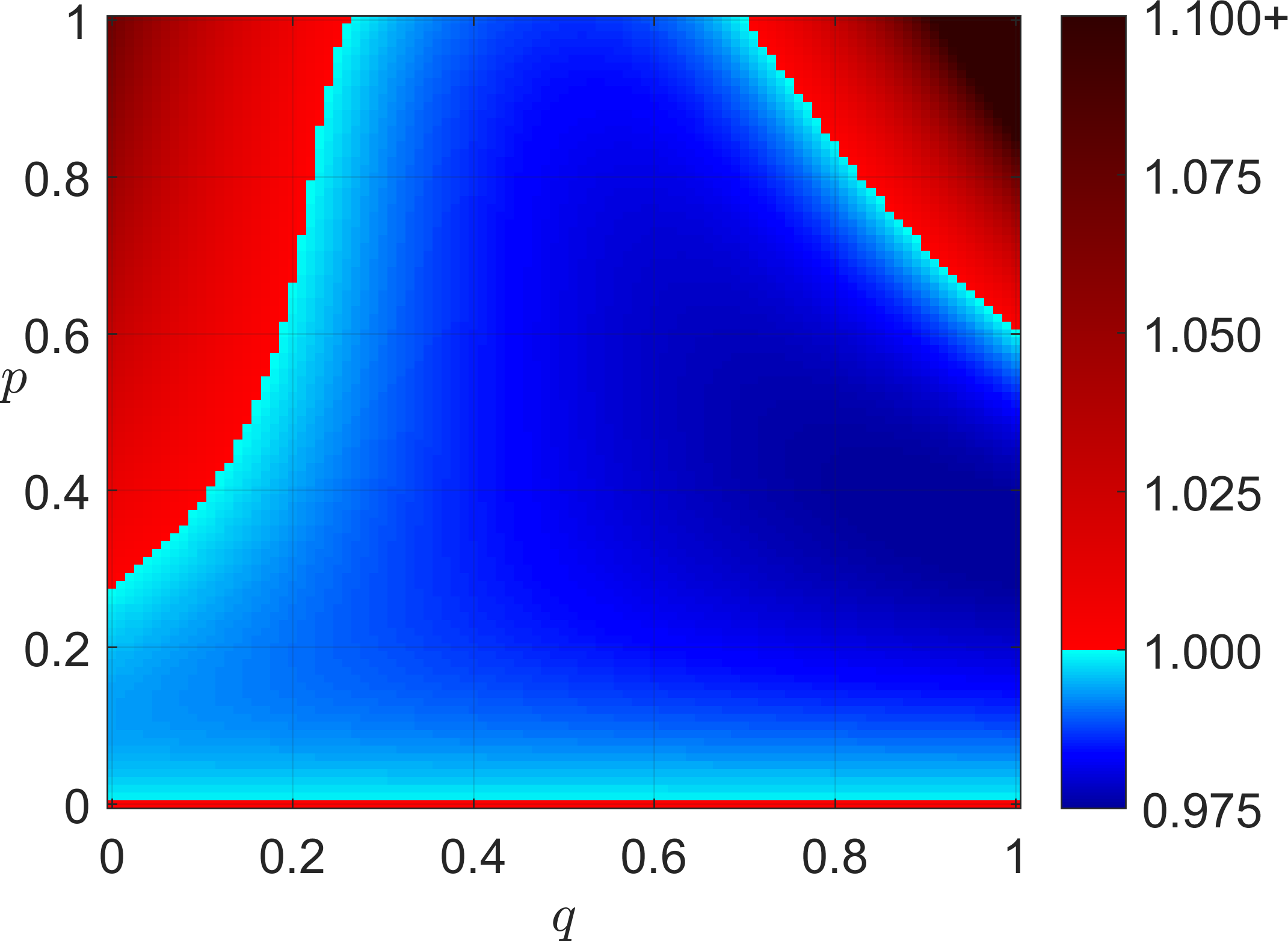}
    \caption*{(a) Mean-square stability of the system with $\AABOLD_1$}
  \end{minipage}
  \hfill
  \begin{minipage}[b]{0.48\textwidth}
    \includegraphics[width=\textwidth]{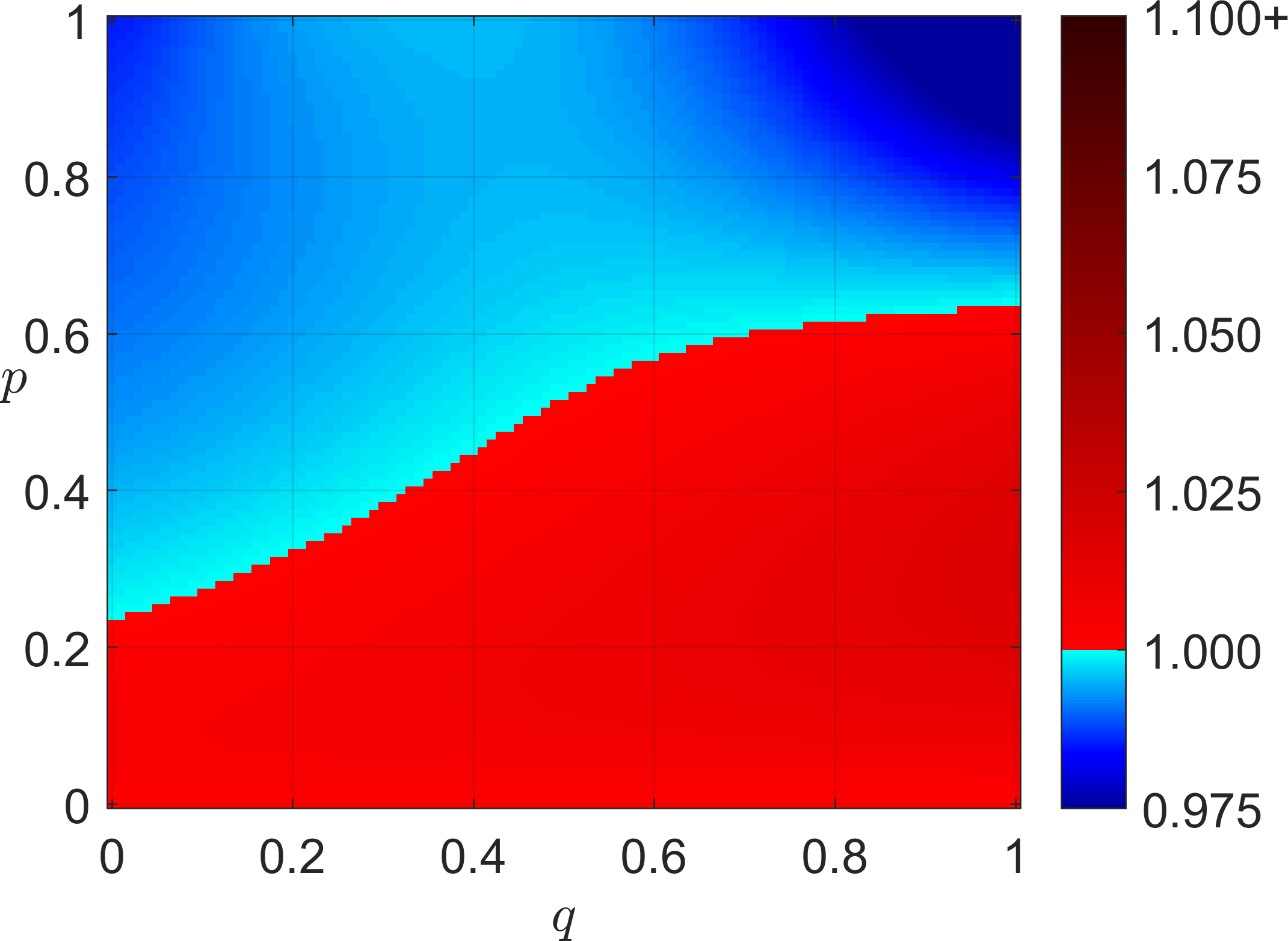}
    \caption*{(b) Mean-square stability of the system with $\AABOLD_2$}
  \end{minipage}
  
  \caption{ The spectral radius of $\vecMap_h$ that represents mean-square stability of the random asynchronous systems with $h=2$ and (a) unstable and (b) stable state transition matrices }
  \label{fig:pq}

\end{figure}

Consider the random asynchronous systems given in (\ref{eq:asyncStateRec}) with the following transition matrices,
\begin{equation} \label{unstable_eg}
    \AABOLD_1 =  \begin{bmatrix} 0.05 & 0.36 & 0.39 \\ 0.01 & -0.37 & 0.23 \\ 
    0.23 & 0.23 & -0.98  \end{bmatrix}, \quad  \AABOLD_2 =  \begin{bmatrix}  0.78 &-1.45 & -1.12 \\ -0.20 & -0.25 & 0.36 \\     0.49 & -0.58 & 0.20 \end{bmatrix}
\end{equation}
with $h=2$. Note that the spectral radius of $\rho(\AABOLD_1) \approx 1.1065$, \textit{i.e.}, the synchronous LTI system with $\AABOLD_1$ is unstable. In order to numerically examine the mean-square stability of the random asynchronous system, we study the system from Markov-jump linear system perspective. In particular, the system in (\ref{eq:asyncStateRec}) can be represented as a Markov-jump linear system of $\dimSta (h+1)$ dimensions that switches between $(1+(h+1)^{\dimSta})^{\dimSta}$ possible systems. Let $\vecMap_h \in \mathbb{R}^{( \dimSta (h+1))^2 \times ( \dimSta (h+1))^2}$ be the matrix that governs the evolution of the correlation matrix of $\dimSta (h+1)$ variables for the Markov-jump system. The stability of $\vecMap_h$ is equivalent to the mean-square stability of the Markov-jump linear system~\citep{markovJumpSystems} and thus the mean-square stability of (\ref{eq:asyncStateRec}). Therefore, for all $p$ and $q$ values, we construct the corresponding $\vecMap_h$ and consider the spectral radius of $\vecMap_h$. Figure \ref{fig:pq}(a) depicts the mean-square stable and unstable regions for all $p$ and $q$ values for the random asynchronous system with $\AABOLD_1$. 

Figure \ref{fig:pq}(a) demonstrates that the synchronous variant ($p=q=1$) of this system is unstable as expected. However, it also shows that by randomizing the updates, \textit{i.e.}, decreasing $p$, or increasing the asynchrony in the updates, \textit{i.e.}, decreasing $q$, one can achieve stable system dynamics from this unstable system. This observation provides a novel perspective to the common perception of randomization and asynchrony. Even though they are usually considered as the ways to decrease the cost or delays in the expense of accuracy and convergence, this result shows that they can be also utilized as the mechanisms to stabilize the dynamical systems. On the other hand, we should note that for this particular system, a drastic increase in asynchrony still results in unstable system dynamics (top left region on Figure \ref{fig:pq}(a)). Moreover, for totally non-random system ($p=1$), moderate levels of asynchrony ($q \in [0.25, 0.65]$) stabilizes the system but we obtain unstable dynamics towards both extremes of asynchrony. This highlights the fact that a careful study is required for understanding the precise effects of randomization and asynchrony on the stability of LTI systems. 

Next, we consider a stable synchronous LTI system with the state transition matrix of $\AABOLD_2$, \textit{i.e.}, $\rho(\AABOLD_2) \approx 0.9778$. Similar to the unstable case, we compute $\vecMap_h$ for this random asynchronous system and Figure \ref{fig:pq}(b) shows the spectral radius of it for all $p$ and $q$ values. The stability behavior of this random asynchronous system is significantly different. Surprisingly, increasing randomization for this system provides mean-square unstable dynamics in all asynchrony conditions. In addition, for moderately non-random updates ($p>0.6$), all levels of asynchrony provides mean-square stability. 

In these settings, the numerical computation of $\vecMap_h$ was feasible since the systems in (\ref{unstable_eg}) have $\dimSta =3$ and $h=2$, yielding $1000$ possible systems to switch between. For large-scale LDS, one needs to compute the closed-form expression of $\vecMap_h$ in order to characterize the mean-square stability of random asynchronous LTI systems. The theoretical analysis of the mean-squared stability of the general system given in (\ref{eq:asyncStateRec}) is left for future work. However, in the following section, we provide the precise closed-form characterization of $\vecMap_h$ for $h=0$ which is a special case of random asynchronous LTI systems where $q=1$ and $0< p \leq 1$, named as randomized LTI systems.  
\newpage
\section{Randomized LTI Systems }

Since $q=1$, randomized LTI systems systems does not have any delays or asynchrony in the system, \textit{i.e.} if the state element getting updated it has access to the most recent information on all states. The random delay probabilities reduces to ${\prob[ k_j = 0]=1}$ for all $j$. Thus, we get the following model:
\begin{align} \label{eq:randomsyncStateRec}
(\xx_{t\textrm{+}1})_i &= \begin{cases}
(\opM \,\xx_{t} + \BB \, \uu_t+ \ww_t )_i, &\quad \textnormal{w.p.} \qquad \pr, \\
(\xx_{t} + \ww_t)_i, &\quad \textnormal{w.p.} \qquad 1 - \pr,
\end{cases} 
\end{align}
This model corresponds to the setting first introduced in  \citet{teke_rand_async}. In the following, we consider the properties of randomized LTI systems.

\subsection{Markov Parameters}
Markov parameters of an LDS is the unique matrix impulse response of the system. For a synchronous LTI system, the Markov parameters of the system ${(\opM, \BB)}$ are given as $\HH_k = \opM^{k-1} \BB$ for $k \geq 1$. From the input-output viewpoint, the randomized updates on the system
with ${(\opM, \BB)}$ can be represented in an average sense as a synchronous LTI system with parameters ${(\randA, \randB)}$ where the average state-transition matrix ${\randA}$ and the average input matrix ${\randB}$ are given as follows:
\begin{equation}
\randA = \pr \, \opM + (1-\pr) \, \II_{\dimSta}, \qquad \randB = \pr \, \BB.
\end{equation}
As a result, Markov parameters of the average system can be obtained as $\randH_k = \randA^{k-1} \randB$, for $k \geq 1$. Notice that Markov parameters of the underlying system and the randomized system can be directly obtained from each other. The $k^{th}$ Markov parameter of the randomized system, ${\randH_k}$, can be written as a linear combination of the first ${k}$ Markov parameters of the synchronous system:
\begin{equation*}
    \randH_k = \left( \pr \opM + (1-\pr) \II_{\dimSta}\right)^{k-1} \pr \BB = \sum\nolimits_{i=1}^{k} \scalebox{1.2}{$\binom{k-1}{i-1}$} \, \pr^{i} (1-\pr)^{k-i} \HH_{i}.
\end{equation*}
More generally, define the first $K$ Markov parameters matrices:
\begin{equation*}
    \markovMat = [\HH_1 \enskip \HH_2 \enskip \cdots \enskip \HH_{K}], \qquad \randmarkovMat = [\randH_1 \enskip \randH_2 \enskip \cdots \enskip \randH_{K}].
\end{equation*}
We have 
\begin{equation}
    \randmarkovMat = \markovMat \,\left( \coefM \otimes \II_{\dimIn}\right),
\end{equation}

where $\coefM \in \mathbb{R}^{K \times  K}$ is an upper triangular matrix with 
\begin{equation*}
    \coefM_{i, j} = \binom{j-1}{i-1}  \pr^{i} (1-\pr)^{j-i}
\end{equation*}
for $j \geq i \geq 1$.
Notice that $\coefM$ does not depend on the system parameters ${(\opM,\BB)}$, and it is determined solely by the probability ${\pr}$. Moreover, ${\coefM}$ has diagonal entries (thus eigenvalues) of ${\coefM_{i,i} = \pr^{i-1}}$. Thus, ${\coefM}$ is always invertible since we trivially assumed that $\pr > 0$. 
This shows that once the average system behavior and the average rate of updates are known, one can idetify the underlying system parameters exactly. When the update probability is ${\pr = 1}$ (synchronous), we get ${\TT = \II_{K}}$ so that ${\randmarkovMat = \markovMat}$. 
Note that the properties above could be trivially extended to measurement feedback systems where instead of exact state information, a noisy linear function of the current state is observed (randomized partially observed LTI systems). 

\subsection{Mean-Squared Stability}
In this section, we precisely characterize the mean-square stability of randomized LTI systems systems which corresponds to the rightmost vertical axes of Figures \ref{fig:pq}(a) and \ref{fig:pq}(b). As discussed above, the dynamics of the randomized LTI system is determined by the matrix ${\randA}$ in an average sense. The stability of the matrix ${\randA}$ is necessary, but not \emph{sufficient} for stability of the system. In order to analyze the mean-square stability, we look for the condition that ensures ${\expc[\xx_t^{}\,\xx_t^{\top}]}$ stays finite as ${t\rightarrow\infty}$. In fact, the steady-state covariance matrix, i.e., ${\lim_{t\rightarrow\infty}\expc[\xx_t^{}\,\xx_t^{\top}] = \xCor}$, can be  found as the solution of the following \emph{extended} Lyapunov equation introduced in \citet{teke_rand_spm}:
\begin{equation}
    \label{eq:errCorrEquation1}
\xCor = \phi(\xCor) + \randB \, \inCorr \, \randB^{\top} + \Big(\frac{1}{p} - 1\Big) \left(\randB \,  \inCorr \, \randB^{\top} \right) \odot \II + \sigma_w^2 \II
\end{equation} 
where the function ${\phi(\cdot)}$ is defined as follows:
\begin{align}
    \phi(\XX)   =   \randA \XX \randA^{\top} \!  +  (\pr - \pr^2)\left( (\opM - \II)  \XX  (\opM^{\top} \! - \II) \right) \odot \II  =  \randA \XX \randA^{\top}   +  \left(\frac{1}{p} - 1\right) \big( (\randA - \II)  \XX  (\randA^{\top} \!- \II) \big) \odot \II \nonumber.
\end{align}
The function ${\phi(\cdot)}$ is a \emph{positive linear map} that controls the evolution of the state covariance matrix in the extended Lyapunov equation. It can be vectorized as $\vect( \phi(\XX) ) = \vecMap \, \vect(\XX)$ where
\begin{equation} \label{eq:vecmap}
     \vecMap = \randA \otimes \randA + (\pr-\pr^2) \, \JJ \, (\opM-\II) \otimes (\opM-\II),
\end{equation}
for $\JJ = \sum_{i=1}^{N} \big( \ee_i^{} \ee_i^{\top} \big) \otimes  \big(\ee_i^{} \ee_i^{\top} \big)$.
${\vecMap\in\mathbb{R}^{\dimSta^2\times \dimSta^2}}$ is the matrix representation of the linear map ${\phi(\cdot)}$ and corresponds to $\vecMap_h$ introduced in Section \ref{sec:RA-LTI} at $h=0$. Note that to extend this to complex valued systems, (\ref{eq:vecmap}) needs element-wise conjugate operations on the left-side of Kronecker products.\footnote{Element-wise conjugation ensures that $\vecMap$ always has a real nonnegative eigenvalue that is equal
to its spectral radius, and the corresponding eigenvector is
the vectorized version of a positive semidefine matrix. This follows from the extensions of the Perron-Frobenius theorem to positive maps in more general settings, 
Theorem 5 of \citet{posOp}.}
Recall that in Section \ref{sec:RA-LTI}, the mean-square stability of the random asynchronous LTI systems is demonstrated numerically in Figures \ref{fig:pq}(a) and \ref{fig:pq}(b). However, with the closed-form expression of $\vecMap$ in (\ref{eq:vecmap}), we can analytically characterize the stability of the randomized LTI system. 
\begin{myLemma} \citep{teke_rand_spm, teke_phdThesis} \label{lem:stabS}
The randomized LTI systems given in (\ref{eq:randomsyncStateRec}) are mean-square stable if and only if ${\rho(\vecMap)<1}$.
\end{myLemma}

The key observation in Lemma~\ref{lem:stabS} is that mean-square stability of the randomized LTI system and the stability of ${\opM}$ do not imply each other, \textit{i.e.}, ${\rho(\vecMap)  <  1}$ and ${\rho(\opM)  <  1}$ are not equivalent in general. Note that Lemma~\ref{lem:stabS} provides the precise characterization of the rightmost axes of Figures~\ref{fig:pq}(a) \& \ref{fig:pq}(b).

\begin{figure}[t]
  \centering
  \includegraphics[width=\textwidth]{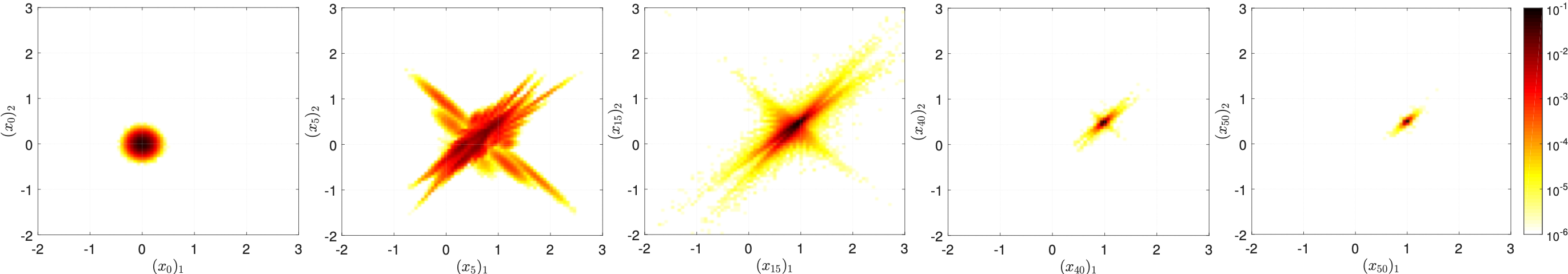}
  \caption{Evolution of the state vector for a mean-square stable (but synchronously unstable) 2-dimensional randomized LTI system with a fixed input and Gaussian initialization}
  \label{fig:mean_square}
\end{figure}

In order to visualize the convergence behavior of the randomized updates, we consider a numerical test example of size ${\dimSta = 2}$ with a constant input (\textit{i.e.}, fixed-point iteration), and initialize ${\xx_0}$ with independent Gaussian random variables (left-most block in Figure~\ref{fig:mean_square}). Then, the distribution of the state vector ${\xx_t}$ (at time ${t}$) follows a Gaussian mixture model (GMM) due to the randomized nature of the updates (See Figure~\ref{fig:mean_square}). Furthermore, the stability of the matrix $\vecMap$ ensures that the mean of ${\xx_t}$ converges to the fixed-point of the system while the variance of ${\xx_t}$ converges to zero.

The key insight to the convergence behavior in Figure~\ref{fig:mean_square} is as follows: When represented as a switching system, the randomized LTI model \eqref{eq:randomsyncStateRec} switches between $2^{\dimSta}$ systems randomly, and it can be shown that all these $2^{\dimSta}$ systems (including the original system) have the same fixed-point. It should also be noted that not all $2^{\dimSta}$ systems are stable by themselves, and an arbitrary switching does not necessarily ensure the convergence. Nevertheless, with a careful selection of the probability, the randomized model can obtain convergence even when the synchronous system is unstable.


When the system is mean-square stable, the steady-state covariance matrix, $\xCor$, is given as
\begin{equation}
	\vect(\xCor) = \left(\II - \vecMap\right)^{-1} \, \left( \left( \pr^2 \, \II + (\pr-\pr^2) \, \JJ \right) \, \vect(\BB \, \inCorr \, \BB^{\top} ) +  \sigma_w^2 \vect(\II) \right).
\end{equation}
When ${\pr=1}$, we have $\phi(\XX) = \opM \, \XX \, \opM^{\top} $, which implies that $\xCor = \opM \, \xCor \, \opM^{\top} + \BB \, \inCorr \, \BB^{\top} + \sigma_w^2 \, \II$ and ${\rho(\vecMap) = \rho^2(\opM)}$. So, we have ${\rho(\vecMap) < 1}$ if and only if ${\rho(\opM)<1}$ for synchronous LTI systems, which recovers the well-known stability result in the classical systems theory.

\section{System Identification for Randomized LTI Systems}

In this section, we propose a system identification method for learning unknown mean-square stable randomized LTI systems from a single input-output trajectory. Regarding the underlying system, we do not have any other assumptions besides stability, \textit{i.e.} $\rho(\vecMap) < 1$, and the assumptions in (\ref{noise_characteristic}). 

First, recall the Markov chain central limit theorem (MC-CLT). Assume that we have a Markov chain at its stationary distribution. MC-CLT states that, the sample average of any measurable, finite-variance and real-valued function of a sequence of $n$ variables from this Markov chain converge to a Gaussian distribution as $n \rightarrow \infty$, where mean is the expected value of this function at the stationary distribution and the variance linearly decays in $n$. 

Notice that the randomized updates of (\ref{eq:randomsyncStateRec}) form an ergodic Markov chain (due to independent selection in every iteration) and the stability of the system guarantees the stationary distribution. We also know that the stable systems converge exponentially fast to their steady state, \textit{i.e.}, Markov chain formed by (\ref{eq:randomsyncStateRec}) quickly approaches to its stationary distribution. In light of these observations, we can deduce that, as the number of collected input-output samples $T$ increases, the sample state correlation and input-output cross correlation matrices converge to their expected values with the rate of $1/\sqrt{T}$. In particular, given a sequence of inputs and outputs $\{\xx_0, \uu_0, \xx_1, \ldots, \uu_{T-1},  \xx_T \}$, let
\begin{equation} \label{sample_correlation}
\CC_0 = \frac{1}{T} \, \sum\nolimits_{t=0}^{T-1} \begin{bmatrix} \xx_{t} \\ \uu_{t} \end{bmatrix} \begin{bmatrix} \xx_{t} \\ \uu_{t} \end{bmatrix}^{\top}, \qquad \CC_1 = \frac{1}{T} \, \sum\nolimits_{t=1}^{T} \, \xx_{t} \begin{bmatrix} \xx_{t-1} \\ \uu_{t-1} \end{bmatrix}^{\top}.
\end{equation}

According to MC-CLT, as $T \rightarrow\infty$, $\CC_0$ and $\CC_1$ converge to $\expc[\CC_0]$ and $\expc[\CC_1]$ respectively, where 
\begin{equation} \label{expectation_correlation}
\expc \left[\CC_0 \right] = \begin{bmatrix} \xCor  &   \bzero   \\ \bzero  &   \inCorr   \end{bmatrix}, \qquad \expc \left[\CC_1\right] = \begin{bmatrix} \randA\xCor & \randB\inCorr \end{bmatrix}.
\end{equation}
Therefore, using $\CC_1 \CC_0^{-1}$ converges to the average state transition and input matrices $\begin{bmatrix} \randA & \randB \end{bmatrix}$. Notice that 
$\CC_1 \CC_0^{-1}$ is in fact the solution of the following least squares problem: 
\begin{equation} \label{leastsquares}
    \arg\min_{\Theta} \sum\nolimits_{t=1}^{T} \tr \Big(\Big(\xx_{t}-\Theta \begin{bmatrix} \xx_{t-1} \\ \uu_{t-1} \end{bmatrix}\Big)\Big(\xx_{t}-\Theta \begin{bmatrix} \xx_{t-1} \\ \uu_{t-1} \end{bmatrix}\Big)^{\top}\Big).
\end{equation}
Thus, we are guaranteed to recover the average system consistently via (\ref{leastsquares}). This result could be extended to recover first $K$ Markov parameters of the randomized partially observable LTI systems. 
Define $\selMat = [\II_{\dimSta} \quad \bzero] \in \mathbb{R}^{\dimSta\times(\dimSta+\dimIn)}$. Then, the extended Lyapunov equation can be written as
\begin{align} \label{eq:errCorrEquation}
\selMat \begin{bmatrix} \xCor  &   \bzero   \\ \bzero  &   \inCorr   \end{bmatrix} \selMat^{\top} \!\! =\!  \begin{bmatrix} \randA~~ \randB \end{bmatrix}  \begin{bmatrix} \xCor  &   \bzero  \\ \bzero  &   \inCorr  \end{bmatrix}   \begin{bmatrix} \randA~~ \randB \end{bmatrix}^{\top}   +  \Big(\frac{1}{p}  -  1\Big) \Big(  \begin{bmatrix} \randA   -  \II ~~ \randB \end{bmatrix}   \begin{bmatrix} \xCor  &   \bzero  \\ \bzero  &   \inCorr   \end{bmatrix}   \begin{bmatrix} \randA   -  \II ~~ \randB \end{bmatrix}^{\top}   \Big)   \odot  \II    
+   \sigma_w^2 \II
\end{align}


We know that covariance matrices of the state variables $\xCor$ and inputs $\inCorr$ must satisfy \eqref{eq:errCorrEquation} for a stable randomized LTI system. The central idea for our system identification method is to exploit this fact and recover the randomization probability $p$, the noise covariance $\sigma_w^2$ and the system parameters $\opM, \BB$ of a stable randomized LTI system. 
Therefore, we can write extended Lyapunov equation \eqref{eq:errCorrEquation} in terms of $\CC_0$ and $\CC_1$ and due to (\ref{expectation_correlation}) expect to have \edlyap$(\CC_0,\CC_1,p, \sigma_w^2) = 0$, where 
\begin{align}
    \edlyap(\CC_0,\CC_1,p, \sigma_w^2) \coloneqq  \selMat \CC_0 \selMat^{\top}   -  \CC_1 \CC_0^{-1} \CC_1^{\top}   -  \Big(\frac{1}{p} - 1 \Big) \Big(  ( \CC_1 \CC_0^{-1}   -  \selMat) \CC_0 ( \CC_1 \CC_0^{-1}   -  \selMat)^{\top} \Big)  \odot \II  - \sigma_w^2 \II. \nonumber
\end{align}

Thus, to identify the underlying system dynamics, we propose to solve the following problem:
\begin{align}
\widehat{p}, \widehat{\sigma}_w^2 &= \arg\min_{\pr,\sigma_w^2} \left\| \edlyap(\CC_0,\CC_1,p, \sigma_w^2)  \right \|_{\fro}^{2}.
\end{align}

This problem can be further simplified to
\begin{equation} \label{lse_p_sig_recovery}
\widehat{\pr}, \widehat{\sigma}_w^2 = \arg\min_{\pr,\sigma_w^2}\left\| \MM_1 - (1/\pr) \, \MM_2 - \sigma_w^2 \, \II \right \|_{\fro}^{2}
\end{equation}
where 
\begin{equation*}
    \MM_2= \Big(  ( \CC_1 \CC_0^{-1}   -  \selMat) \CC_0 ( \CC_1 \CC_0^{-1}   -  \selMat)^{\top} \Big) \odot \II, \qquad \MM_1 = \selMat \, \CC_0 \, \selMat^{\top} - \CC_1^{} \, \CC_0^{-1} \, \CC_1^{\top} + \MM_2.
\end{equation*}
Notice that $\pr$ and $\sigma_w^2$ appear decoupled in (\ref{lse_p_sig_recovery}). Therefore, we can first solve (\ref{lse_p_sig_recovery}) for ${\widehat{\sigma}_w^2}$ for a fixed value of $\pr$ to get an optimal solution. Then, substituting ${\widehat{\sigma}_w^2}$ into the problem and solving for $\widehat{\pr}$ we obtain the optimal estimate for $p$. The described procedure yields the following optimal estimates:

\begin{equation}
\widehat{\pr} = \frac{\dimSta \, \tr(\MM_2^{\top}\,\MM_2)-\tr^{2}(\MM_2)}{\dimSta\tr(\MM_1^{\top}\,\MM_2)-\tr(\MM_1)\tr(\MM_2)}, \qquad \widehat{\sigma}_w^2  = \frac{\tr(\MM_1) - (1/\widehat{\pr}) \, \tr(\MM_2) }{\dimSta}.
\end{equation}

Using the estimate of randomization probability $\widehat{\pr}$ and $\CC_1 \CC_0^{-1} = [\widehat{\randA}~\widehat{\randB}]$, \textit{i.e.}, the estimate of average system transition parameters, the underlying system parameters could be recovered as 
\begin{equation}
    \widehat{\opM}  =  (1/\widehat{\pr})\widehat{\randA}  +  (1 - 1/\widehat{\pr}) \II_{\dimSta} \qquad \text{and} \qquad \widehat{\BB}  =  (1/\widehat{\pr}) \widehat{\randB}.
\end{equation}

\begin{figure}[t]
  \centering
  \begin{minipage}[b]{0.48\textwidth}
  \centering
    \includegraphics[width=\textwidth]{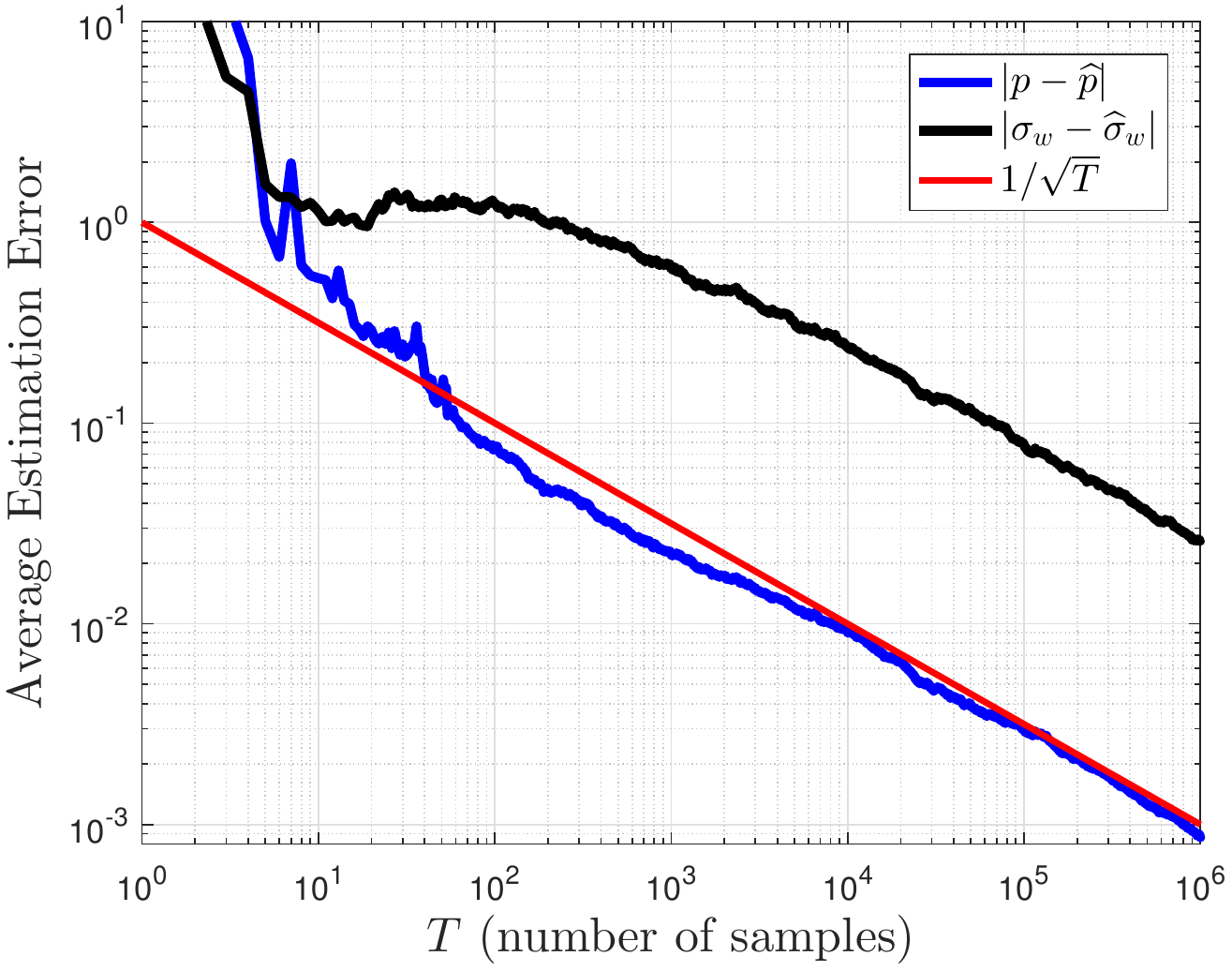}
    \caption*{(a) Average estimation error rate for $p$ and $\sigma_w^2$}
     \label{fig:prob_sig}
  \end{minipage}
  \hfill
  \begin{minipage}[b]{0.48\textwidth}
    \includegraphics[width=\textwidth]{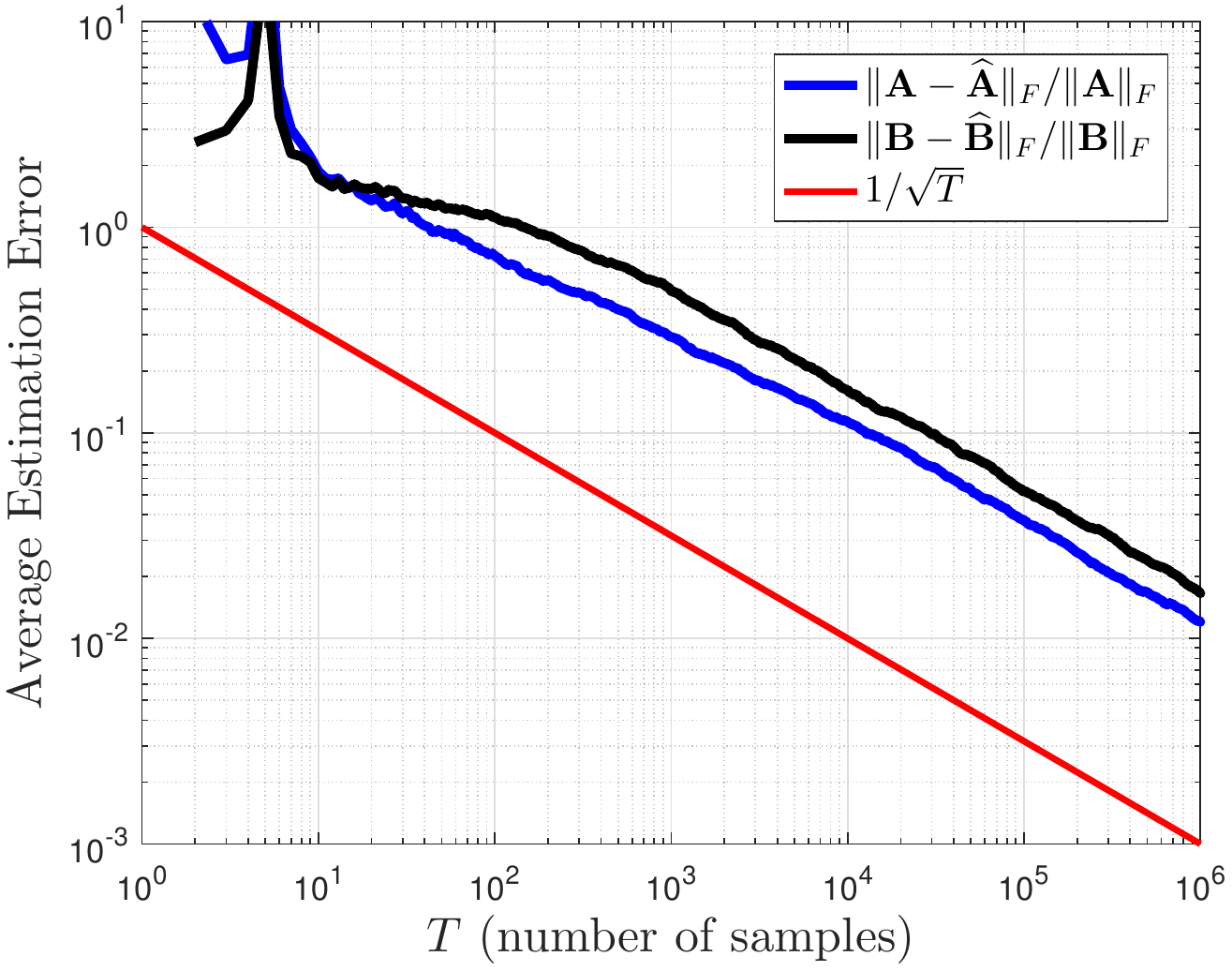}
    \caption*{(b) Average estimation error rate for $\AABOLD$ and $\BB$}
     \label{fig:AB}
  \end{minipage}
  \caption{ Average estimation error for the unknown system parameters of the stable randomized LTI system with state transition matrix of $\AABOLD_1$ and random $\BB$ for 100 independent single trajectories}
  \label{fig:estimation}
\end{figure}

To study the performance of the proposed system identification method, we consider a randomized LTI system with state transition matrix of $\AABOLD_1$ and a random $\BB$ with $\pr = 0.5$ which guarantees the stability (verified by Lemma \ref{lem:stabS}). We run 100 independent single trajectories and present the average rate of decay for the estimation errors of $\pr$, $\sigma_w^2$ in Figure \ref{fig:estimation}(a) and $\opM$ and $\BB$ in Figure \ref{fig:estimation}(b).

Notice that the estimation errors behave irregularly at the beginning where there are few samples, corresponding to burn-in period to converge to steady-state. On the other hand, Figures  \ref{fig:estimation}(a) \& \ref{fig:estimation}(b) show that, as predicted by MC-CLT, the estimation errors decay with $1/\sqrt{T}$ rate as we get more samples. This estimation error rate is the optimal behavior in linear regression problems with independent noise and covariates \citep{hastie2009elements}. This depicts the consistency and efficiency of the proposed system identification method for randomized LTI systems.

\section{Conclusion}
In this work, we introduced a natural model of random asynchronous LTI systems which can be used in modeling various LDS that have randomized and asynchronous updates. We numerically and analytically studied the mean-square stability of these systems and showed that the stability of random asynchronous systems is governed by the matrix representation of a positive linear map that controls the evolution of the state covariance matrix, rather than state transition matrix. We provided a consistent and efficient system identification method for stable randomized LTI systems. 

In future work, we aim to derive the precise characterization of mean-square stability and extend the proposed system identification method in the general random asynchronous systems. We also plan to study the finite-time adaptive control and stabilization of random asynchronous LTI systems. 


\bibliographystyle{plainnat}
\bibliography{allRefs}

\end{document}